\documentclass{article}

\usepackage[nonatbib, final]{neurips_2019}

\usepackage[utf8]{inputenc} 
\usepackage[T1]{fontenc}    
\usepackage{url}            
\usepackage{booktabs}       
\usepackage{amsfonts}       
\usepackage{nicefrac}       
\usepackage{microtype}      

\usepackage{enumitem}   

\usepackage{dsfont}
\usepackage{xspace}

\usepackage[dvipsnames,usenames]{xcolor}
\definecolor{NavyBlue}{cmyk}{0.94,0.54,0,0}
\definecolor{DarkBlue}{rgb}{0,0.2,0.6}

\usepackage[colorlinks=true, linkcolor=BrickRed,citecolor=DarkBlue,urlcolor=DarkBlue,bookmarks=false]{hyperref}

\usepackage{cleveref} 

\usepackage[pdftex]{graphicx}

\usepackage{bm}          
\usepackage{nicefrac}

\usepackage{amssymb,amsmath,amsthm, mathtools}
\usepackage{amscd}
\usepackage{float}    

\usepackage{caption, subcaption}

\usepackage{siunitx} 

\usepackage{xspace}

\usepackage[
    backend=bibtex,
    style=numeric,
    natbib=true,    
    url=false, 
	isbn=false,
    doi=false,
	maxnames=2, 
	maxbibnames=8, 
	uniquename=false,
    eprint=true
]{biblatex}
\usepackage{wrapfig}

\usepackage[ruled,vlined]{algorithm2e}

 \usepackage[colorinlistoftodos]{todonotes}

\newtheorem{lemma*}{Lemma}

\def\R{\mathbb{R}}

\def\x{\mathbf{x}}

\def\0{\mathbf{0}}

\newcommand{\suchthat}{\;\ifnum\currentgrouptype=16 \middle\fi|\;}

\DeclarePairedDelimiterX{\inner}[2]{\langle}{\rangle}{#1, #2}

\DeclareMathOperator*{\argmax}{argmax}

\DeclarePairedDelimiterX{\infdivx}[2]{(}{)}{%
  #1\;\delimsize\|\;#2%
}

\newenvironment{itemize*}%
  {\begin{itemize}%
  \vspace{-0.5cm}
    \setlength{\itemsep}{0pt}%
    \setlength{\parskip}{0pt}}%
  {\end{itemize}}
\newenvironment{enumerate*}%
{\begin{enumerate}
    \vspace{-0.5cm}
    \setlength{\itemsep}{0pt}%
    \setlength{\parskip}{0pt}}%
  {\end{enumerate}}

  {\list{}{\leftmargin=#1\rightmargin=#1}\item[]}%
  {\endlist}

\def\woe{\text{woe}}

\usepackage{cleveref}
\crefname{section}{§}{§§}
\Crefname{section}{§}{§§}

\title{Weight of Evidence as a Basis \\ for Human-Oriented Explanations}

\author{%
  David Alvarez-Melis \\
  Microsoft Research\\
  \hspace*{0cm}\texttt{alvarez.melis@microsoft.com}\hspace*{0cm}\\
  \And
  Hal Daum\'e III  \\
  Microsoft Research \& University of Maryland \\
  \texttt{me@hal3.name}\\
   \AND
   Jennifer Wortman Vaughan  \\
   Microsoft Research \\
    \hspace*{1cm}\texttt{jenn@microsoft.com}\hspace*{1cm}\\
    \And 
   Hanna Wallach \\
   Microsoft Research \\
   \hspace*{1.3cm}\texttt{wallach@microsoft.com}\hspace*{1.3cm}\\
}

\begin{document}

\maketitle

\begin{abstract}
    Interpretability is an elusive but highly sought-after characteristic of modern machine learning methods. Recent work has focused on interpretability via \textit{explanations}, which justify individual model predictions. In this work, we take a step towards reconciling machine explanations with those that humans produce and prefer by taking inspiration from the study of explanation in philosophy, cognitive science, and the social sciences. We identify key aspects in which these human explanations differ from current machine explanations, distill them into a list of desiderata, and formalize them into a framework via the notion of \textit{weight of evidence} from information theory. Finally, we instantiate this framework in two simple applications and show it produces intuitive and comprehensible explanations.
\end{abstract}

\section{Introduction}    
With the growing success of complex predictors, and their resulting expanding reach into high-stakes and decision-critical applications, wringing explanations out of these models has become a central problem in artificial intelligence (AI). Countless methods have been recently proposed to produce such explanations \citep{simonyan2013deep, selvaraju2016grad, Ribeiro2016Why, lundberg2017unified, alvarez-melis2017causal}, yet there is no consensus on what precisely makes an explanation of an algorithmic prediction good or useful. Meanwhile, what it means to \emph{explain} and how humans do it are questions that have been long studied in philosophy and cognitive science. Since the end goal of explainable AI is to explain \textit{to humans}, this literature seems an appropriate starting point when looking for principles upon which a theory of \textit{machine} interpretability might rest.

While the debate on the nature of (human) explanation is far from settled, various fundamental principles arise across theoretical frameworks. For example, at the core of Van Fraassen's \citep{vanfraassen1988pragmatic} and Lipton's \citep{lipton1990contrastive} theories of explanation is the hypothesis that we tend to explain in \textit{contrastive} terms  (e.g., ``fever is more consistent with pneumonia than with a common cold"), focusing on both factual and counterfactual explanations (e.g., ``had this patient had chest pressure too, the diagnosis would have instead been bronchitis"). On the other hand, both of Hempel's models of explanation \citep{hempel1962deductive} are characterized by sequences of simple premises, reflecting the fact that humans usually explain using multiple simple accumulative statements, each one addressing a few aspects of the evidence (e.g., ``presence of fever rules out cold in favor of bronchitis or pneumonia, and among these two, the presence of chills suggests the latter"). These and other fundamental principles have been observed across disciplines in the social sciences. In a recent survey of over 250 papers in philosophy, psychology and cognitive science on explanation, \citet{miller2019explanation} mentions \textit{contrastiveness} and \textit{selectivity} (i.e., that only few possible cases are presented) as two major properties of the way humans explain things that he argues are important for explainable AI but yet are currently under-appreciated.

These principles are often missing in popular explainable AI frameworks. Their explanations consist of saliency or attribution scores that are \textit{absolute} (i.e., non-contrastive, focused only on the predicted outcome), purely \emph{factual} (i.e., based only on aspects present in the input, ignoring counter-factuals), and \textit{monolithic} (i.e., simultaneously explicating all input features). On the other hand, those that provide probabilistic explanations often present only posterior probabilities, which conflate class priors (also known as base rates) with per-class likelihoods, and which humans are notoriously bad at reasoning about \citep{tversky1974judgment, bar-hillel1980base, eddy1982probabilistic, koehler1996base}. Furthermore, as Miller \citep{miller2019explanation} and others argue, attribution is only an important but incomplete \textit{part} of the entire process of human explanation. This novel view of explanation as a \textit{process} rather than (only) a \textit{product}, is crucial for understanding the discrepancy between current approaches to automated interpretability and the way humans explain. 

In this work, we lay out a general framework for interpretability that aims to reconcile this discrepancy. The starting point of this approach, and our first contribution, is a set of intuitive desiderata that we argue are crucial for bringing machine explanations closer to their human counterparts. With these considerations at hand, we develop a mathematical framework to realize them. At the core of this framework is the concept of \textit{weight of evidence} from information theory, which we show provides a suitable theoretical foundation to the often elusive notion of model interpretability.\footnote{While the use of weight of evidence for algorithmic explainability has previously been advocated by David Spiegelhalter (e.g., in his keynote talk at NeurIPS 2018 \citep{spiegelhalter2018neurips}), to the best of our knowledge it has not yet been instantiated or investigated in the context of complex machine learning models.} After introducing this concept, we extend it beyond its original formulation to account for the type of settings in machine learning where interpretability is most needed (e.g., high-dimensional, multi-class prediction). We provide a generic meta-algorithm to produce explanations based on the weight of evidence, and show its instantiation on simple proof-of-point experimental settings.

\vspace{-0.3cm}
\paragraph{Related Work} Some of the shortcomings of machine explanations highlighted here have been individually tackled in prior work. For example, recent work seeks to move from absolute to contrastive or counterfactual explanations \citep{wachter2017counterfactual, miller2018contrastive, vanderwaa2018contrastive}, partly inspired by earlier approaches on contrast set mining \citep{Azevedo2010Rules, Bay1999Detecting, webb2003detecting, Novak2009Supervised}. On the other hand, while most saliency-based methods produce dense high-dimensional attributions, explanations supported on a sparse set of input features is a much-touted benefit of classic (model-based) interpretability, such as decision trees \citep{quinlan1986induction} and sets \citep{Lakkaraju2016Interpretable}. Recent work has also explored improving interpretability by explaining on higher-level concepts (e.g., super-pixels or patterns in an image) rather than raw inputs \citep{kim2018interpretability, alvarez-melis2018towards}. Our approach shares motivation but many of these works, but differs substantially in how the salient features are selected and scored.


\section{Desiderata for Human-Oriented Explanations}
The first step towards defining any method for explainable machine learning should be to define its goal with precision, i.e., what is an explanation? For this, we draw on basic principles and terminology from epistemology and philosophy of science. In its most abstract form, an explanation is an answer to a \textit{why-question} \citep{hempel1948studies, vanfraassen1988pragmatic} consisting of two main components: the \textit{explanandum}, the description of a phenomenon to be explained; and the \textit{explanans}, that which gives the explanation of the phenomenon \citep{hempel1948studies}. Different ways to formalize the explanans have given rise to various theories of explanation; an excellent historical overview of these can be found surveys by \citet{pitt1988theories} and \citet{miller2019explanation}. For the purposes of this work, our definition of \textit{interpretability} follows that of \citet{biran2017explanation} and \citet{miller2019explanation}: the degree to which an observer can understand the cause of a decision. 

In the context of machine learning, we are usually interested in explanations of predictive models. For a predictor $f :\mathcal{X} \rightarrow \mathcal{Y}$ that takes inputs $x \in \mathcal{X}$ drawn according to some distribution $\x\sim D$ and produces outputs $y\ \in \mathcal{Y}$, we seek an explanation for $y=f(x)$, that is, "\textit{why did model $f$ predict $y$ on input $x$}?" Here, we are primarily interested in probabilistic---or more generally, soft---predictors, which covers a wide range of machine learning methods. Namely, we consider models that rather than producing a single prediction $y$, instead return a predictive posterior distribution $P( Y \suchthat X=x)$. Furthermore, to allow for more general explananda (e.g., \textit{why was this subset of outcomes ruled out?}), we take inspiration from hypothesis testing and consider complex hypotheses of the form $h:  y \in U \subseteq \mathcal{Y}$, and---slightly abusing notation---denote the posterior as $P( h \suchthat X = x)$.

Having described what type of explanandum we consider in this work, we must now characterize the explanans we seek. The first such consideration pertains to the \textit{causes} or \textit{evidence} which are to define the ``vocabulary'' from which the explanans is constructed. Popular explanation-based interpretability methods rely directly on the raw inputs $x$. Likewise, we initially consider evidence $e$ of the form $e =  \{x_i\}_{i \in V}, V \subseteq \{1,\dots,n\}$, but will later generalize to more general \textit{attributes} (e.g., subsets of features). Extending this definition to include higher-level representations of the input \citep{kim2018interpretability, alvarez-melis2018towards} or even aspects of the model itself (e.g., parameters) are natural extensions that we leave for future work.

Having formalized its ingredients, we now discuss what properties the explanans should have. Recall that our objective is to devise machine interpretability methods that are intelligible to humans. Our survey of literature on explanations above highlighted various aspects that characterize human explanations, but which most current machine explanations lack. Based on these, we propose a set of \textit{desiderata} for bridging the gap between the former and the latter. Namely, explanations should:

\begin{enumerate}[leftmargin=1.5em,nolistsep,noitemsep]
    \item \textbf{be contrastive}, i.e., answer the question ``why did model $f$ predict $y$ \textit{instead of} $y'$?". 
    \item \textbf{be modular and compositional}, which is particularly important whenever the relations between inputs and outputs/predictions are complex -- precisely when interpretability is most needed.
    \item \textbf{not confound base rates with input likelihood}, i.e., while important for fully understanding a classifier, base rates should be presented separately from input relevance towards the predictions.
    \item \textbf{be exhaustive}, i.e., they should explicate why every other alternative $y'$ was not predicted.
    \item  \textbf{be minimal}, i.e., all things being equal the simpler of two explanations should be preferred. 
\end{enumerate}

Next, we propose an interpretability framework based on the weight of evidence\;---a basic but fundamental concept from information theory---\;that satisfies all of the desiderata above. 

\section{Explaining with the Weight of Evidence}
\label{sec:background}
\subsection{Weight of Evidence: from Information Theory to Bayesian Statistics}

The weight of evidence (WoE) is an information-theoretic approach to analyze variable effects in prediction models \citep{good1950probability, good1968corroboration, good1985weight}. Although originally defined in terms of log-odds (see supplement), the weight of evidence for a hypothesis $h$ in the presence of evidence $e$ can be conveniently defined as $\woe(h : e) \triangleq \log \frac{P(e \mid h)}{P(e\mid \overline{h})}$. The interpretation of this quantity is simple. If $\text{woe}(h:e) > 0 $ then $h$ is more likely under $e$ than marginally, i.e., the evidence \textit{speaks in favor of hypothesis} $h$. Analogously, $\text{woe}(h:e) < 0 $ indicates $h$ is less likely when taking into account the evidence than without it.

The WoE can be conditioned on additional information:  $\text{woe}(h : e \mid c) \triangleq  \log \frac{P( e \mid h, c) }{P(e \mid \overline{h}, c)}$, and can be computed relative to an arbitrary alternative hypothesis (i.e., not necessarily the complement): $\text{woe}(h/h': e) \triangleq  \text{woe}(h: e \mid h \lor h')$. Thus, we can in general talk about the evidence in favor of $h$ and against $h'$ provided by $e$ (and perhaps conditioned on $c$). Further properties and an axiomatic derivation of WoE are provided in the supplement. An appealing aspect of the WoE is its immediate connection to Bayes' rule. For this, consider the binary classification setting, i.e., $h: Y=1$, $\bar{h}: Y = 0$ and $e= X$. Simple algebraic manipulation of the definition of WoE yields:
\begin{equation}\label{eq:woe_alternative}
    \log \frac{P(Y=1 \mid X)}{P(Y=0 \mid X)} = \underbrace{\log\frac{P(Y=1)}{P(Y=0)}}_{\text{Sample log-odds}} + \underbrace{\log \frac{P(X \mid Y=1)}{P(X \mid Y=0)}}_{\textit{Weight of evidence}}.
\end{equation}
This provides another useful interpretation of the WoE in classification: a positive (negative, resp.) WoE implies that the posterior log-odds (of $Y=1$ over $Y=0$) are higher (lower) than the base log-odds, showing that $X$---the evidence---speaks in favor of (against) the hypothesis $h: Y=1$.

Besides being intuitive and well-understood, the WoE provides an appealing framework for machine interpretability because it immediately satisfies three of the interpretability desiderata introduced in the previous section: it is naturally contrastive ($\text{woe}(h/h': e)$ quantifies the evidence in favor of $h$ \textit{against} $h'$), it decouples base log-odds from variable importance (Eq.~\eqref{eq:woe_alternative}) and it admits a modular decomposition (Eq.~\eqref{eq:woe_multievidence} in the supplement). We later show how the last two desiderata can be met.

\subsection{Sequential Explanations: Explaining High-Dimensional Multi-Class Classifiers}\label{sec:extension}

The weight of evidence has been mostly used in simple settings, such as a single binary outcome variable $Y$ and a single input variable $X$. Its use in the (typically more complex) settings considered in machine learning poses various challenges. First, in multi-class classification one must choose the contrast hypotheses $h$ and $h'$. The trivial choice of letting $h$ be the predicted class $c^*$ and $h'$ its complement is unlikely to yield interpretable explanations when the number of classes is very large (e.g., explaining the evidence in favor of one disease against 999 other possibilities). To address this, we take inspiration from Hempel's model \citep{hempel1962deductive} and propose to cast explanation as a sequential process, whereby a subset of the possible outcomes is \textit{expounded away} in each step. For example, in medical diagnosis this could correspond to first explaining why bacterial diseases were ruled out in favor of viral ones, then contrasting between viral families, and finally between the predicted disease and similar alternatives. In general, we consider explanantia consisting of $q+1$ nested hypotheses $h_0 : \{ Y = c^* \} \subseteq h_1 \subseteq \dots \subseteq h_q = V$, which imply $q$ contrastive tests $h_{i-1}/h_{i}$.

A second challenge in using WoE for complex prediction tasks arises from the size of the input. While the decomposition formula (Eq.~\eqref{eq:woe_multievidence}) allows us to produce individual scores for each feature, for high-dimensional inputs (such as in images or detailed health records), providing a WoE score for every single feature simultaneously will rarely be informative. Thus, we propose grouping the inputs into \textit{attributes} (e.g., super-pixels for images or groups of related symptoms for medical diagnosis). Formally, we partition the set of $n$ input features into $m$ subsets: $S_1 \cup \dots \cup S_m = \{1,\dots,n\}$. 

Given these two extensions of the WoE, we propose a simple meta-algorithm for generating explanations for classifiers. At every step, a subset $C_t$ of the classes is selected to keep (the rest are \textit{ruled out}), and $\woe(C_{t}/\overline{C_{t}} : X)$ is computed using the decomposition formula \eqref{eq:woe_multievidence}. The user is presented with only the most relevant attributes (cf.~desideratum 5) according to their WoE (e.g., using the rule-of-thumb threshold of $\pm 2$ \citep{good1985weight}), in addition to the base log-odds $\log P(C_t)/P(\overline{C_t})$. This process continues until all classes except the predicted one $c^*$ have been "ruled out" (desideratum 4). It is important to note that unless the predictor is generative---and not black-box---this process requires \textit{estimating} the conditionals $P(X_{S_i} | Y)$. A discussion on estimation, in addition to pseudo-code for this method (Algo.~\ref{algo:greedy_woe}) and details about its implementation are provided in the supplement.               

\section{Experiments}\label{sec:experiments}
\begin{figure}
    \centering
    \includegraphics[width=0.43\linewidth]{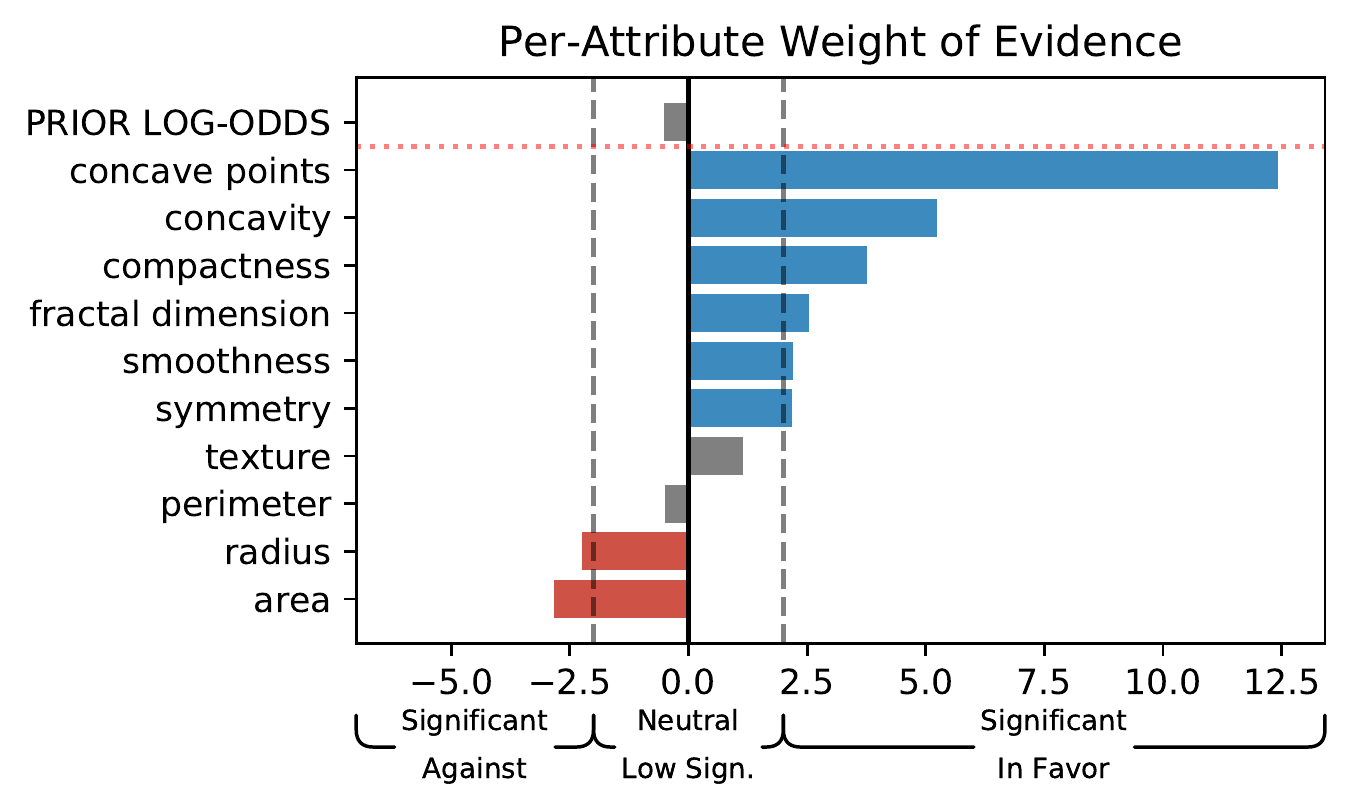}%
    \includegraphics[width=0.23\linewidth, angle=90]{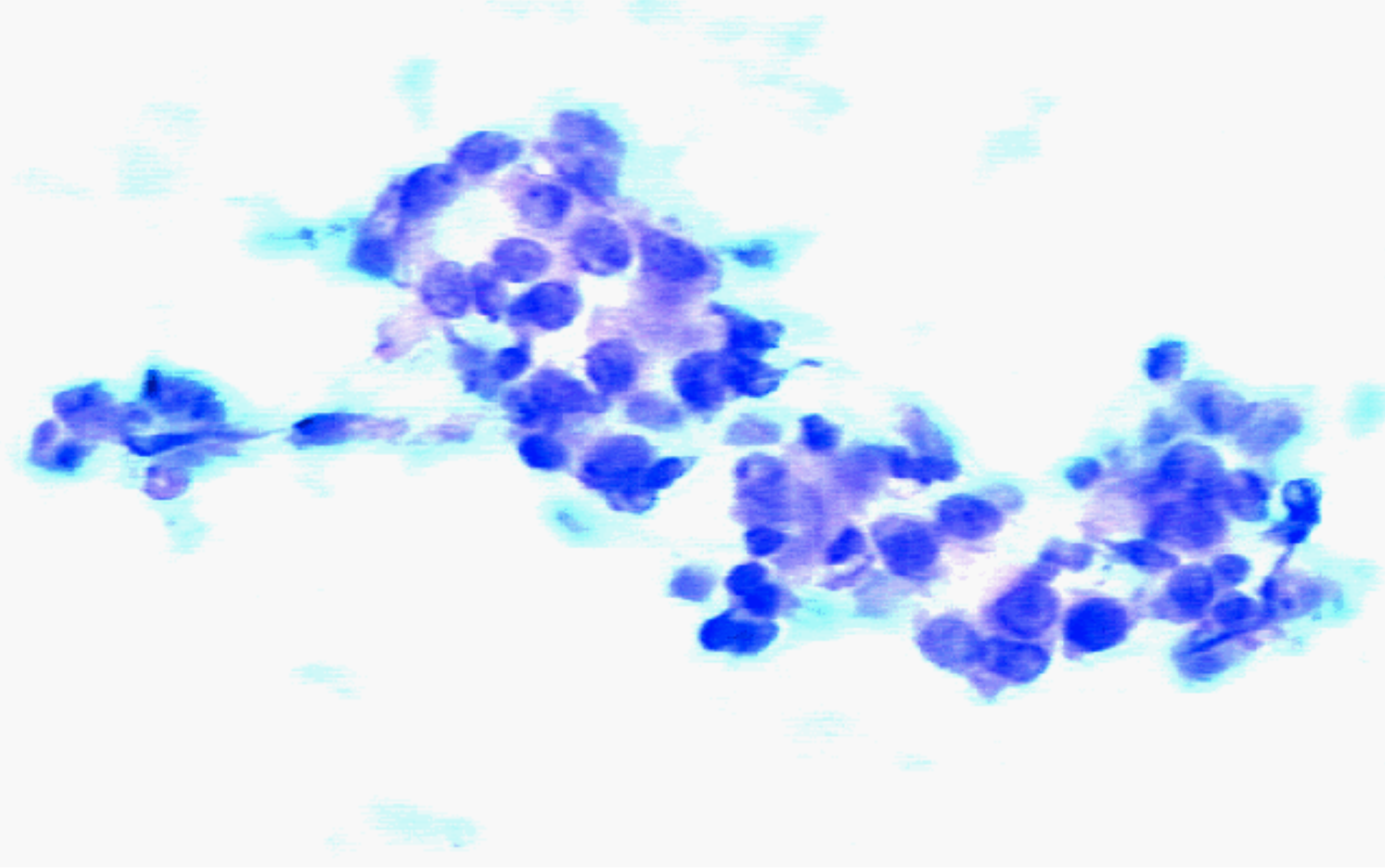}\hfill%
    \includegraphics[width=0.4\linewidth]{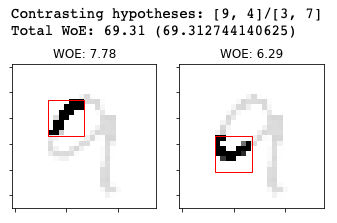}%
    \caption{\textbf{Left}: A WoE explanation (in favor of \texttt{malign}) in the Breast Cancer dataset for a test example (original tissue image shown). \textbf{Right}: MNIST classification. The attributes shown (others given in supplement) have high WoE for explaining the prediction of classes [\texttt{9},\texttt{4}] against [\texttt{3},\texttt{7}].}
    \label{fig:examples}
\end{figure}
We first illustrate our framework in a simplified setting with \textit{exact} WoE computation (i.e., without estimation) using a Gaussian Naive Bayes classifier, which intrinsically computes $P(X|Y)$ as part of its prediction rule. We use the Wisconsin Breast Cancer dataset, grouping the 30 scalar-valued features into 10 attributes according to their type (mean/s.e./worst area$\rightarrow$ \texttt{area}, etc.). In the example in Fig.~\ref{fig:examples} (left), the model predicts \texttt{malignant} despite initial log-odds, \texttt{radius}, and \texttt{area} speaking slightly against it, because the cell's \texttt{concavity} and \texttt{compactness} attributes speak very strongly in favor of malignancy. 

In our second experiment, we use our framework to explain the predictions of a black-box neural-net MNIST classifier, estimating conditional probabilities via a masked autoregressive flow (MAF) model \citep{papamakarios2017masked}, using squared $7\times 7$ super-pixels as attributes. For the example explanation in Fig.~\ref{fig:examples} (right), the strong evidence in favor of classes \texttt{9},\texttt{4} (against \texttt{3} or \texttt{7}) clearly corresponds to parts of the image which would be uncharacteristic for examples of the latter classes.
\section{Discussion and Extensions}


We have proposed a set of desiderata for bridging the gap between the type of explanations provided by humans and current interpretability methods, and a promising framework to realize them based on the weight of evidence. The application of this concept to complex machine learning problems brings about various challenges, some of which we addressed here (high-dimensional inputs and multi-class classification), but many which remain, such as estimation of WoE scores for black-box models, selection of contrast hypotheses, and attribute design. Furthermore, since the ultimate beneficiary of these explanations is a human, the effect of the proposed solutions to these challenges\;---and all algorithmic choices---\;should be validated and compared through human evaluation. 



\printbibliography

\pagebreak
\clearpage

\appendix

\section{Desiderata for Interpretability in Further Detail}

Our driving motivation in this work is to devise machine interpretability methods that emulate the way humans explain. In the introduction, we listed some aspects that characterize human explanations (and which most current \textit{machine} explanations lack). Based on these, we proposed in Section 2 a set of \textit{desiderata} for interpretabilty which are aimed at bridging the gap between the former and the latter. We discuss them here in much more detail.

\vspace{-0.3cm}
\paragraph{D1. Explanations should be contrastive.} As mentioned before, several authors have proposed (and validated) that humans tend to explain in contrastive terms. To more faithfully emulate human cognition, machine explanations should be contrastive too. That is, the prototypical explanandum should not be ``why did model $f$ predict $y$?'', but rather, ``why did model $f$ predict $y$ instead of $y'$?''. Despite how self-evident this might be, note that most current explanation methods are not contrastive. Instead, they explain the model's prediction \textit{absolutely}, leaving the contrast case undetermined or implicitly assuming it to be the complement of the predicted class.\footnote{For example, an explanation for a prediction of "9" by a digit classifier is to be interpreted as "why 9 and not any other digit"?}

\vspace{-0.3cm}
\paragraph{D2. Explanations should be modular and compositional.} Interpretabilty is most needed in applications where the inputs, outputs or the causal relations between them are complex (and therefore so is any interesting statistical model whose goal is to predict these). Yet, in these settings most explainable AI methods produce a single, high-dimensional static explanation for any given prediction (e.g., a heatmap for an image classifier). These are often hard to analyze and draw conclusions from, particularly for non-expert users. In addition, this form factor again differs from the manner in which humans tend to explain \citep{hempel1962deductive}: using various simple premises. Thus, instead of a single monolithic explanans, we seek a set of \textit{simple} sub-clauses, each explicating a different aspect of the input-output predictive phenomenon. Clearly, this modularity introduces a trade-off between the number of clauses in the explanans (too many clauses might be difficult to coherently analyze simultaneously) and their relative complexity (small clauses are easier to reason about, but more of these might be required to explain a complex predictor). At a higher level, breaking up an explanation into a sequence of small components responds to our goals of moving from \textit{explanation as a product} towards \textit{ explanation as a process} \citep{lombrozo2012explanation} and to emulate the selective aspect of human explanation \citep{hilton2017social, miller2019explanation}.

\vspace{-0.3cm}
\paragraph{D3. Explanations should not confound base rates with input likelihoods.} When explaining probabilistic models, any human-oriented framework for interpretability should take into account how humans understand and interpret probabilities.  The psychological and cognitive science communities have long studied this topic \citep{tversky1974judgment}, showing, for example, that humans are notoriously bad at incorporating class priors when thinking about probabilities. The classic example of Breast Cancer diagnosis due to  \citet{eddy1982probabilistic}, showed that the majority of subjects (doctors) tended to provide estimates of posterior probabilities roughly one order of magnitude higher that the true values. This phenomenon has been attributed to a neglect of base-rates during reasoning (the \emph{base-rate fallacy} \citep{bar-hillel1980base}), or instead, to a confusion of inverse conditional probabilities $P(A|B)$ and $P(B|A)$, one of which needs to be estimated and the other one is provided (the \emph{inverse fallacy}, \citep{koehler1996base}). Whatever the cause, we argue here that its effect---i.e., that humans often struggle to reason about posterior probabilities---should be taken into account. Thus, explanations should clearly separate the contribution of base-rates and per-class likelihoods linking inputs and predictions. We argue that while both of these are important for understanding a prediction, their very different natures (one of them dependent on the input of the other one not) necessitates different treatment. To the best of our knowledge, no currently available off-the-shelf interpretability framework provides this.

\vspace{-0.3cm}
\paragraph{D4. Explanations should be exhaustive.} A conclusive justification for a predicted hypothesis $h$ should explicate why no other alternative hypothesis was predicted. In Hempel's terminology, we seek explanations that are \textit{complete} (not to be confused with \textit{complete} in the sense of \citet{goodman2006intuitive}, i.e., where all \textit{variables} are explained). For example, an explanation for a pneumonia diagnosis simply stating the presence of cough is non-exhaustive, since cough by itself could be indicative of various other conditions beyond pneumonia, so their being non-predicted should be justified.

\vspace{-0.3cm}
\paragraph{D5. Explanations should be minimal.} In following the original purpose of Occam's Razor, if two explanations are of different complexity but otherwise identical (in particular, both sufficiently explicate the prediction), the simpler of these should be preferred. Furthermore, if omitting less-relevant aspects of an explanation makes the whole more intelligible, while remaining equally faithful to the prediction being explained, then then trimmed explanation should be preferred. 

\section{The Weight of Evidence: Properties and Axiomatic Derivation}

The weight of evidence is a fundamental concept that has been introduced in many contexts\footnote{The basic principle behind the weight of evidence appear in the work of both Alan Turing and Claude Shannon. However, \citet{good1985weight} claims ideas like it go back to at least \citet{peirce1878probability}).}, although it is primarily associated with I.J. Good who popularized it through a long sequence of works \citep{good1950probability, good1968corroboration, good1985weight}. Good originally defined it as follows. For a hypothesis $h$ in the presence of evidence $e$, the weight of evidence of $h$ is defined as
\begin{equation}\label{eq:woe_def}
    \text{woe}(h : e) \triangleq  \log \frac{O( h \mid e) }{O(h)}
\end{equation}
where $O(\cdot)$ are the log-odds, i.e.,
\begin{equation}
    O(h) := \frac{p(h)}{p(\bar{h})}, \qquad O( h \mid e) := \frac{p(h \mid e)}{p(\bar{h} \mid e)}
\end{equation}
The interpretation of \eqref{eq:woe_def} is simple. If $\text{woe}(h:e) > 0 $ then $h$ is more likely under $e$ than marginally, i.e., the evidence \textit{speaks in favor of hypothesis} $h$. Analogously, $\text{woe}(h:e) < 0 $ indicates $h$ is less likely when taking into account the evidence than without it.

The WoE has various desirable theoretical properties. For example, \citet{good1985weight} provides an axiomatic derivation for Definition \ref{eq:woe_def}, showing that it is (up to a constant) the only function $F$ of $e$ and $h$ that satisfies the following properties:
\begin{enumerate}
    \item $F$ is a function of the likelihoods, i.e., $F[p(e\mid h), p(e\mid \bar{h})]$
    \item The posterior is a function of the prior and $F(e,h)$, i.e.,  $p(h\mid e) = g [p(h), F(h,e)]$
    \item $F$ is additive (on the evidence) (indeed, $\text{woe}(h : e_1 \land e_2) = \text{woe}(h :e_1) + \text{woe} (h: e_2 \mid e_1)$)
\end{enumerate}

The following two properties, which are easy to prove, are crucial for our extension into complex models in Section~\ref{sec:extension}:
\begin{align}
    \woe(h/h' : e) &= \log \frac{P(e \mid h)}{P(e\mid h')} \\ 
    \woe(h/h': e_1 \land e_2 \land \dots \land e_n) &= \sum_{i=1}^n \log \frac{P(e_i \mid e_{i-1}, \dots, e_1, h)}{P(e_i \mid e_{i-1}, \dots, e_1, h')}    \label{eq:woe_multievidence}
\end{align}
The first of these provides a simple expression to compute WoE scores. The second one will prove consequential to defining an intelligible extension of WoE to high dimensional inputs.

\section{Explanations of Complex Models via the Weight of Evidence}

As mentioned in the main text, using the WoE framework for complex machine learning models brings about the challenge of keeping WoE scores interpretable despite (i) high-dimensional inputs and (ii) not necessarily binary output (e.g., multi-class classification) settings. 

We address (ii) by sequentially contrasting (increasigly smaller) sets of classes $C_i$ (i.e., complex hypotheses), as described in the main text. Our solution for (i) in turn involves grouping the inputs into \textit{attributes}, e.g., super-pixels in an image or groups of related symptoms in our running medical diagnosis example. Consider an input space of dimension $n$, i.e., the evidence $\mathbf{e}$ corresponds now to a multivariate random variable $X$ taking values in $\R^n$. Property \eqref{eq:woe_multievidence} (or alternatively, the chain rule of probability) allows for chaining of the conditional probabilities; hence, for any a partition of the $n$ input features into $m$ attributes, we can express the WoE of hypothesis $h: \{ y \in C\}$ against $h': \{y \in C'\}$ as:
\begin{equation}\label{eq:woe_attributes}
   \text{woe}( h/h' : \mathbf{e}) =   \sum_{i=1}^m \underbrace{\log \frac{P(X_{S_i} \mid  X_{S_{i-1}}, \dots, X_{S_1}, Y \in C)}{P(X_{S_i} \mid X_{S_{i-1}}, \dots, X_{S_1}, Y\in C')}}_{\triangleq \woe(C/C': X_{S_i} \mid  X_{S_{i-1}}, \dots, X_{S_1})}
\end{equation}
where $X_{S_i}$ denotes the subset of random variables with indices in $S_i$, i.e., $X_{S_i}=\{X_j\}_{j \in S_i}$. The full Bayes-odds explanation model now has the form 
\begin{equation}\label{eq:woe_final}
\underbrace{\log \frac{P(Y \in C \mid X)}{P(Y \in \bar{C} \mid X)}}_{\text{posterior log-odds of entailed set}} = \underbrace{\log\frac{P(Y \in C)}{P(Y \in \bar{C})}}_{\text{prior log-odds of entailed set}} + \sum_{i=1}^m \underbrace{\woe(C/C': X_{S_i} \mid  X_{S_{i-1}}, \dots, X_{S_1})}_{\text{conditional WoE of i-th attribute }}
\end{equation}
Note that, in general, the order of the attributes matters in this sum. We discuss how to minimize the impact of this ordering in the next section. These two extensions lead to the meta-algorithm for WoE-based explanation shown here as Algorithm~\ref{algo:greedy_woe}, which selects at each step the contrast set $C_i$ with maximal WoE plus a cardinality-based regularizing term $R(U)$ to prevent too small or too large subsets from being selected. We use $R(U) = \alpha(|U| - \tfrac{1}{2}|V|)^{2}$ to encourage even partitions. 

\begin{algorithm}[t]
\SetAlgoLined
\SetKw{KwParam}{Parameters}
\KwIn{Example $X \in \R^n$, class $c^* \in \{1,\dots, K\}$ predicted by the model.}
\KwParam{Attribute size $\alpha$, hypothesis size regularization $\lambda$.}

 $V \gets \{1,\dots, K\}$ \tcp*{Remaining classes to be explained}
 \While{$|V| > 1$}{
  \tcc{Find hypothesis maximizing regularized `total' woe}
  $U \gets \argmax_{U \subseteq V; c^* \in U} \woe(U/(V\setminus U) : X) - R(U)$\;
  $\tilde{U} \gets V \setminus U$ \tcp*{contrast hypothesis is the relative complement} 
  \tcc{Compute base log-odds of chosen hypothesis}
  $\text{lod}(U) \gets \log \tfrac{P(U\mid X)}{P(\tilde{U} \mid X)}$\;
  $T \gets \{1\, \dots, n\}$\;
  $i \gets 0$\;
  \While{$ |T| > \alpha$}{
    \tcc{Find attribute with maximal partial 
    woe}
    $S_i \gets \argmax_{S \subset T; |S| = \alpha} \woe(U/\tilde{U}) : X_S)  $\;
    $\omega_i \gets \woe(U/\tilde{U} : X_{S_i})  $\;
    $T \gets T \setminus S_i$
  }
  $\text{DisplayExplanation}(U, \tilde{U}, \{S_i\}, \text{lod}(U), \{\omega_i\})$\;
  
  $V \gets V \setminus U$\tcc*{Update classes to be explained}
  }
 \caption{Weight of evidence explanation generation for complex models}\label{algo:greedy_woe}
\end{algorithm}

\section{Estimation of Weight of Evidence Scores}\label{sec:estimation}

For any pair of entailed and contrast classes, and any partition of input into attributes, equation \eqref{eq:woe_multievidence} provides an exact method to compute the conditional weight of evidence as a sum of per-attribute WoE scores. In order to use this expression, we need to be able to compute $P(X_{S_i} \mid X_{S_{i-1}}, \dots, X_{S_1}, Y \in C)$ for any order of attributes $\{X_{S_i}\}_{i=1}^n$ and outcome set $C \subset \mathcal{Y}$. In an ideal scenario (such as the Gaussian Naive Bayes classifier used in Section~\ref{sec:experiments}, or in an auto-regressive generative model for sequential data), the prediction model itself would compute and store these values. 

Unfortunately, most prediction models do not compute such probabilities explicitly, so any realistic application of the WoE methodology to interpretability must provide a fallback method to estimate these, independently, from data. Let us consider the worst case scenario: a black-box prediction model, for which we assume we only have oracle access (i.e., queries of $f(x)$), in addition to access to additional training data. In such case, the problem essentially turns to a conditional density estimation problem, where we seek to learn models of $P(X_{S_i} | Y)$ from training data in the form of pairs ($x, \hat{y}$), where $x$ is a sample from the input distribution $X \sim P_X$ and $\hat{y} = f(x)$ is a class label prediction obtained with the prediction model. 

We propose to tackle this problem by training, as an off-line preliminary step, an auto-regressive conditional likelihood estimation model. For simple data, this could be done with classic (e.g., kernel or spectral) density estimation methods. For more complex data such as images, many recent methods have been proposed based on normalizing flows and autoregressive models \citep{rezende2015variational, dinh2016density, papamakarios2017masked}. For sequential data, the ordering of the attributes in Eq.~\eqref{eq:woe_attributes} is implied by the data. For non-sequential inputs, the likelihood model should be trained to minimize the impact of ordering on the WoE scores (e.g., by training on random orderings).  For the experiments on MNIST, we train a conditional Masked Autoregressive Flow (MAF) model \citep{papamakarios2017masked}, randomizing the order in which the $7\times 7$ pixel blocks (the attributes) are traversed, but keeping the order within each of these fixed (left-to-right, top-to-bottom). 

\section{A full explanation of the Black-Box MNIST classifier}

\includegraphics[width=0.95\linewidth]{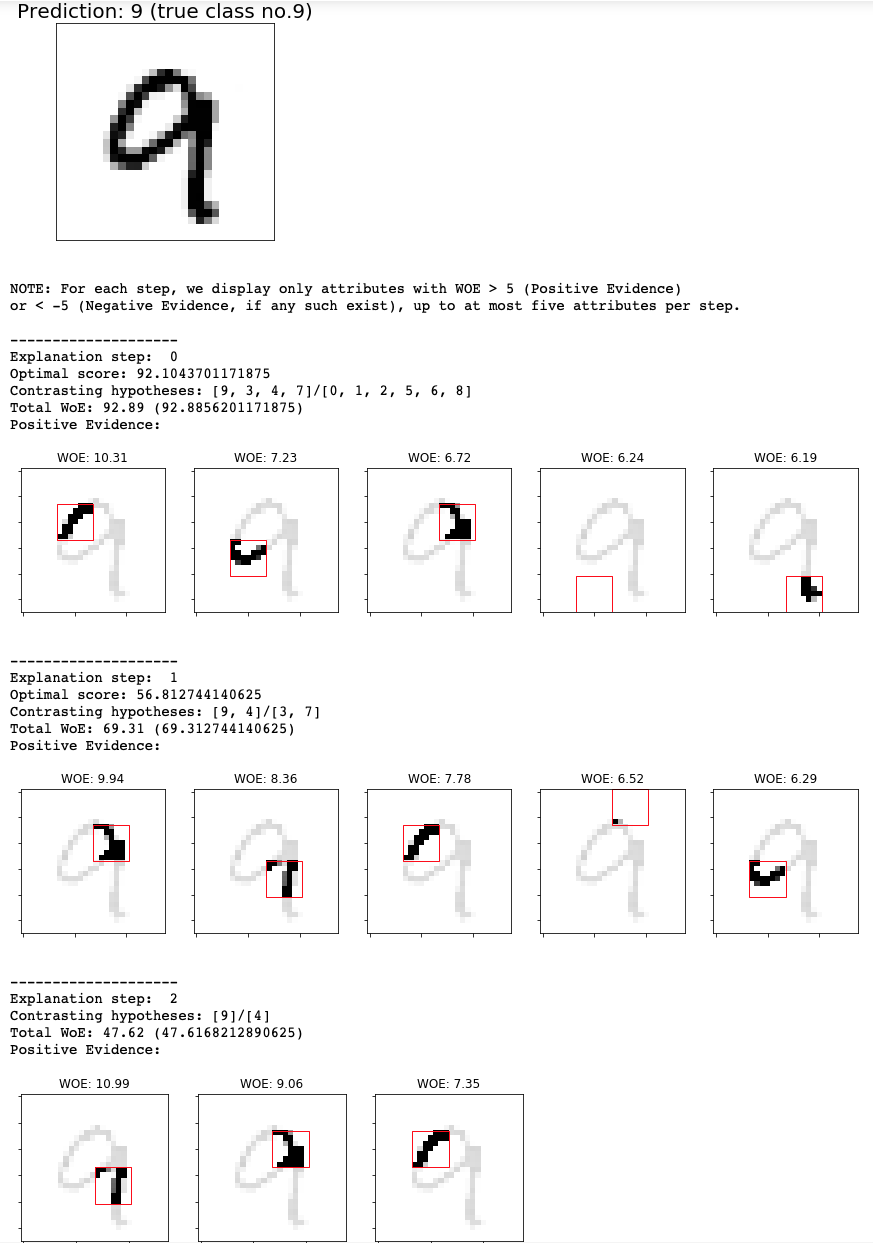}

\end{document}